\documentclass[letterpaper, 10 pt, conference]{ieeeconf}  

\IEEEoverridecommandlockouts                              

\usepackage{graphics} 
\usepackage{graphicx}
\usepackage{times} 
\usepackage{amsmath} 
\usepackage{amssymb}  
\usepackage{mathrsfs}
\usepackage{amsfonts}
\usepackage{kantlipsum}
\usepackage{booktabs}

\makeatletter
\let\NAT@parse\undefined
\makeatother
\usepackage[numbers,sort&compress]{natbib}

\usepackage{hyperref}
\usepackage{graphicx}
\usepackage{float}
\usepackage{ctable}
\usepackage{cuted}
\usepackage{colortbl}
\usepackage{multirow}
\let\labelindent\relax
\usepackage{enumitem}
\usepackage{amsmath}
\usepackage{amssymb}
\usepackage{algpseudocode}
\usepackage{algorithm}

\usepackage{marvosym}
\usepackage{graphicx}
\usepackage{float}
\usepackage{ctable}
\usepackage{cuted}
\usepackage{colortbl}
\usepackage{multirow}
\usepackage{threeparttable}
\usepackage{makecell}
\usepackage{wrapfig}
\usepackage{threeparttable}

\definecolor{green}{RGB}{0,150,10}
\definecolor{blue}{RGB}{0,148,181}
\definecolor{orange}{RGB}{194,153,107}

\title{\LARGE \bf
UniAff: A Unified Representation of Affordances for Tool Usage and Articulation with Vision-Language Models
}
\author{Qiaojun Yu$^{1*}$, Siyuan Huang$^{1,7*}$, Xibin Yuan$^{1}$,  Zhengkai Jiang$^{2}$, Ce Hao$^{3}$, \\ Xin Li$^{1}$, Haonan Chang$^{4}$, Junbo Wang$^{1}$, Liu Liu$^{5}$, Hongsheng Li$^{6}$, Peng Gao$^{7}\textsuperscript{\Letter}$ and Cewu Lu$^{1}\textsuperscript{\Letter}$ 
\thanks{$^{1}$Qiaojun Yu, Siyuan Huang, Xibin Yuan, Xin Li, Junbo Wang, Cewu Lu are with the Shanghai Jiao Tong University, China. $^2$Zhengkai Jiang is with Hong Kong University of Science and Technology, HongKong. $^3$Ce Hao is with National University of Singapore, Singapore. $^4$Haonan Chang is with Rutgers University, United States of America. $^5$Liu Liu is with Hefei University of Technology, China.$^6$Hongsheng Li is with CUHK-MMLab, China. $^7$Siyuan Huang and Peng Gao are with the Shanghai AI Lab, China. * indicates the equal contribution. \Letter Peng Gao and Cewu Lu are the equal corresponding authors, \texttt{gaopeng@pjlab.org.cn, lucewu@sjtu.edu.cn}.}
\thanks{\textbf{Acknowledgement}. This work was supported in part by the National Natural Science Foundation of China under Grant (62302143), the Anhui Provincial Natural Science Foundation under Grant (2308085QF207), the National Key Research and Development Project of China (2022ZD0160102), the Shanghai Artificial Intelligence Laboratory, and XPLORER PRIZE grants (2021ZD0110704).}%
}

\begin{document}

\maketitle

\begin{abstract}

Previous studies on robotic manipulation are based on a limited understanding of the underlying 3D motion constraints and affordances. To address these challenges, we propose a comprehensive paradigm, termed \textit{UniAff}, that integrates 3D object-centric manipulation and task understanding in a unified formulation. Specifically, we constructed a dataset labeled with manipulation-related key attributes, comprising 900 articulated objects from 19 categories and 600 tools from 12 categories. Furthermore, we leverage MLLMs to infer object-centric representations for manipulation tasks, including affordance recognition and reasoning about 3D motion constraints. Comprehensive experiments in both simulation and real-world settings indicate that UniAff significantly improves the generalization of robotic manipulation for tools and articulated objects. We hope that UniAff will serve as a general baseline for unified robotic manipulation tasks in the future. Images, videos, dataset and code are published on the project website at:https://sites.google.com/view/uni-aff/home.

\end{abstract}


\section{Introduction}
Mastering the manipulation of tools and articulated objects is crucial for embodied robots, which requires understanding physical constraints and interaction regions in 3D space~\cite{kaelbling2020foundation, geng2023sage}. For effective manipulation, identifying movable parts, joint types, 3D joint parameters, and affordances in articulated objects is essential~\cite{geng2023gapartnet, yu2024gamma}. Similarly, modeling 6D pose, grasping regions and functional areas is vital for tool use in specific tasks~\cite{li2024learning, huang2024copa}. By integrating task-related 6D pose, 3D motion constraints, and affordances predictions, embodied robots can further enhance their adaptability and efficiency.

\begin{figure}[t]
    \centering
\includegraphics[width=0.9\columnwidth]{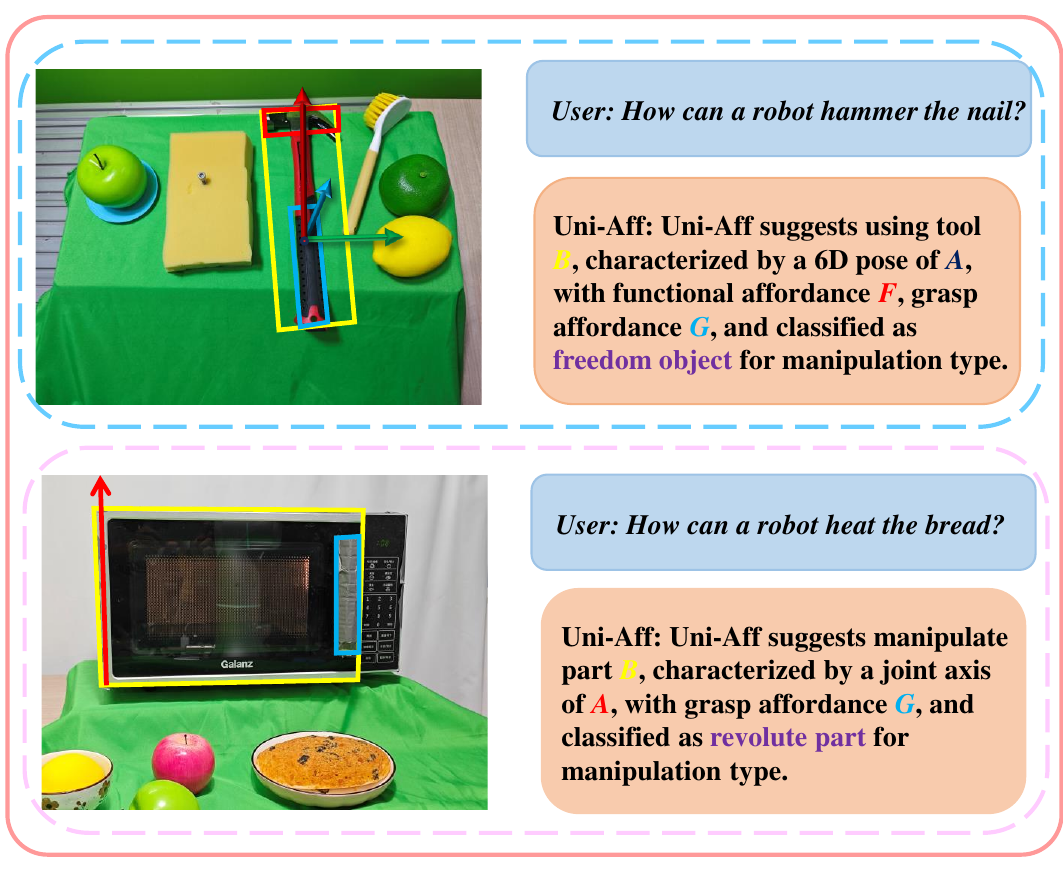}
    \vspace{-0.5cm}
    \caption{UniAff demonstrates its ability to unify tool usage and articulation understanding in a VQA format, predicting part bounding boxes, 6D poses, grasp affordances, functional affordances, and manipulation types, etc for effective robotic manipulation tasks.}
    \label{Fig: teaser}
\vspace{-4mm}
\end{figure}

Most existing approaches focus exclusively on either articulated objects~\cite{huang2024a3vlm, yu2024gamma,li2024manipllm} or tools~\cite{huang2024manipvqa, guo2023handal, myers2015affordance}, which limits their ability to generalize across tasks. By leveraging the reasoning capabilities of large language models (LLMs), two-stage methods are employed: the first stage predicts manipulation-related parameters using a vision model~\cite{xia2024kinematic, geng2023sage}, while the second stage utilizes the LLM's reasoning. Additionally, they address case-by-case problems without offering general task reasoning capabilities.  To overcome these limitations, we propose UniAff, a unified representation of affordances for both tools and articulated objects, powered by vision-language models, as illustrated in Figure~\ref{Fig: teaser}. UniAff combines object-centric manipulation with task understanding, utilizing multimodal large language models (MLLMs) to improve comprehension of 3D motion constraints and affordances.

For the training of UniAff, We developed a comprehensive dataset for articulated object manipulation and tool-use tasks, including 900 articulated objects over 19 categories and 600 tools in 12 categories. Each object's part-level 6D pose, grasp affordances, manipulation types, and functional affordances are labeled, forming a multifunctional dataset for robot learning. By unifying the formulation of manipulation tasks, we incorporate object-centric 3D motion constraints and affordances. Near-realistic simulations generate pre-scanned meshes or URDF models~\cite{tola2024understanding}, efficiently producing a large-scale dataset with automatically labeled information.

To equip the MLLM with unified affordance capabilities, we fine-tune the SPHINX model~\cite{lin2023sphinx} on our dataset, enabling UniAff to predict object-centric 3D representations and infer affordances for both tools and articulated objects. To our best knowledge, UniAff is the first model offering a unified understanding of both categories, representing a significant advancement in object-centric robotic manipulation.

Extensive experiments in both simulation and real-world settings demonstrate UniAff's effectiveness. On the HANDAL dataset~\cite{guo2023handal}, UniAff outperformed LISA~\cite{lai2024lisa} by $11.5\%$ and closely matched ManipVQA~\cite{huang2024manipvqa}, with only a $2.2\%$ IOU difference, showcasing strong adaptation to real-world tasks. For articulated object manipulation, UniAff improved the success rate by $7.07\%$ on unseen instances and $9.60\%$ on unseen categories compared to A3VLM~\cite{huang2024a3vlm}.
These results demonstrate UniAff’s ability to generalize and adapt across tasks. In summary, our contributions are:
\begin{enumerate}
    \item Introducing UniAff, an MLLM that facilitates understanding of fine-grained physical properties, 3D motion constraints, and affordances for manipulation.
    \item Developing a dataset with 1,500 objects across 19 categories of articulated objects and 12 categories of tools, labeled with part-level 6D poses, manipulation types and affordances.
    \item Conducting comprehensive experiments that demonstrate UniAff's significant improvement in robotic manipulation generalization across articulated objects, and tools.
\end{enumerate}

\section{Related Work} \label{Sec: related works}


\textbf{Datasets for Object-centric Manipulation.} 
Object-centric manipulation focuses on understanding object poses and affordances for various tasks. By focusing on object representation, object-centric policies are more transferable across robots, but they also require high-quality, diverse data on object poses and affordances. For affordance data, the RGB-D Part Affordance dataset~\cite{myers2015affordance} provides part-level labels for 105 tools, while PhysObjects~\cite{gao2024physically} offers annotations of household objects, focusing on physical properties. In object pose estimation, Omni6DPose~\cite{zhang2024omni6dpose} provides extensive pose annotations for real and simulated images. HANDAL~\cite{guo2023handal}, an early attempt to combine object affordance and pose estimation, includes graspable regions but is limited to 210 objects and only grasp affordances. To address these limitations, UniAff introduces a large-scale synthetic dataset, utilizing 230 real-world scanned tools from the PACE~\cite{you2023pace} and OmniObject3D~\cite{wu2023omniobject3d} datasets, articulated data from PartNet-Mobility dataset~\cite{Xiang_2020_SAPIEN}, along with 370 newly scanned tools. Additionally, we include 900 articulated object manipulations across 19 categories and 600 tool-use tasks across 12 categories.

arios.**

\textbf{Instruction-based Robotic Manipulation.} Instruction-based robotic manipulation connects abstract commands to low-level control. Early works like CLIPort~\cite{shridhar2022cliport} and 6D-CLIPort~\cite{zheng2022vlmbench} utilize pre-trained text encoders, such as CLIP~\cite{radford2021learning}, for task-specific policies. Hiveformer~\cite{guhur2023instruction} integrates language, observations, and action history, while Perceiver Actor~\cite{shridhar2023perceiver} uses voxelized 3D data for efficient learning. However, pre-trained language encoders are limited to simple instructions and lack deeper reasoning for complex tasks. To address these limitations, LLMs have been applied to task planning~\cite{wu2023tidybot, cai2023bridging, chang2023lgmcts}, robot control code generation~\cite{liang2023code, huang2023voxposer} and VLA~\cite{song2024germ, li2024robonurse}.  Recent advances, such as RT-2-X~\cite{padalkar2023open}, ManipLLM~\cite{li2024manipllm}, and AIC-MLLM~\cite{xiongautonomous}, incorporate multimodal LLMs (MLLMs) like LLaVA~\cite{liu2024visual} and Sphinx~\cite{gao2024sphinx}, improving reasoning from both language instructions and visual input. However, these methods rely on modeling robot actions directly, requiring extensive real-world interaction data, which limits transferability across different robots. To overcome this, works like ManipVQA~\cite{huang2024manipvqa} and A3VLM~\cite{huang2024a3vlm} emphasize affordance reasoning and articulation awareness. Yet, ManipVQA focuses only on tool use, and A3VLM on articulated objects, restricting their real-world applications. To address this, we propose UniAff, a unified representation for both tools and articulated objects.


\section{Method}

In this section, we introduce our method, which presents a structured 3D spatial formulation for object representation in robotic manipulation. Our approach unifies task formulation by incorporating object-centric 3D spatial motion constraints and their corresponding affordances (Section~\ref{Subsec: task formulation}). To further enhance spatial intelligence, we developed a synthetic dataset that applies this unified formulation to both tools and articulated objects (Section~\ref{Subsec: data gen}). Finally, we fine-tune MLLMs to integrate object-centric formulations through Visual Question Answering (VQA) (Section~\ref{Subsec: model training}). 

\begin{figure*}
    \centering
    \includegraphics[width=0.95\textwidth]{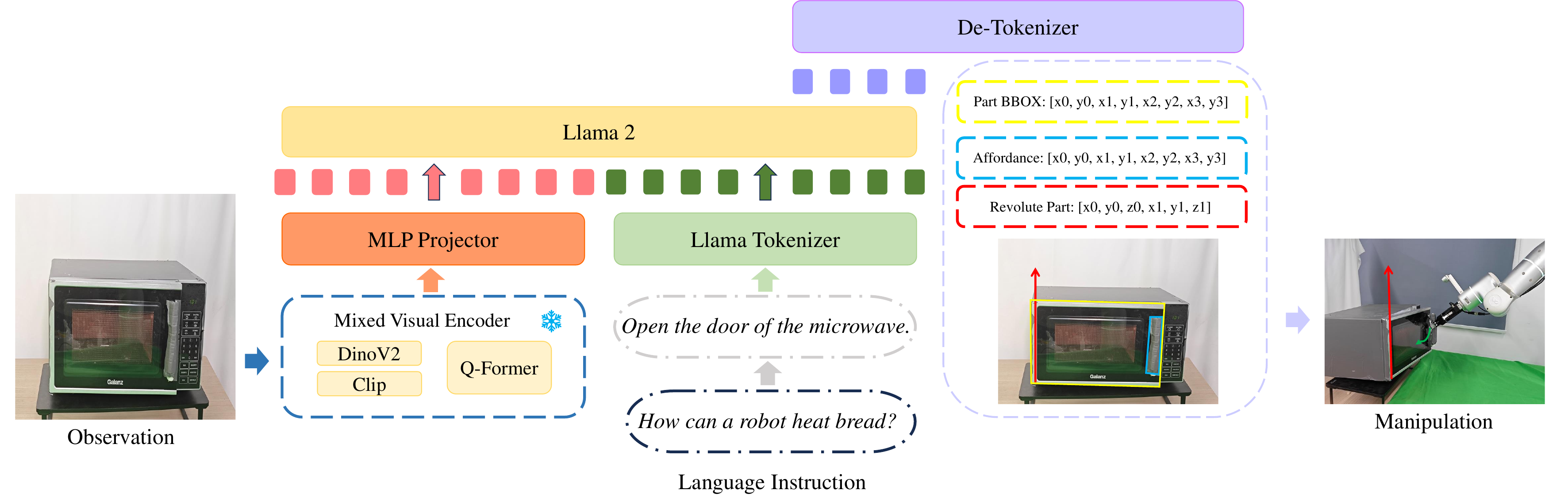}
      \caption{\textbf{The architecture of UniAff}. The image features are first extracted using a Mixed Visual Encoder, such as DINOv2, CLIP, or Q-Former, followed by an MLP projector. Next, language instructions are used to extract features with the Llama Tokenizer. Finally, the output of the structured manipulation tasks, such as Part BBOX, Affordance, and Revolute Parts, is used to execute robotic instructions.}
    \label{Fig: pipeline}
\vspace{-4mm}
\end{figure*}

\subsection{Formulation of Structured Manipulation Task} \label{Subsec: task formulation}

We define the manipulation task formulation $T$ as follows.
An unknown object $M$ consists of $K$ movable parts, represented as $M=\{m_{i}\}_{i=1}^{K}$. We observe the object $M$ through an image $I$ and a depth map $D$. Furthermore, we define the object structure $S$ for each part $\psi_i$ as $S=\{\psi_{i}\}_{i=1}^K$. The parameters for each part can be defined as $\psi_i=\{\mathcal{A}_i, \mathcal{B}_i, \mathcal{G}_i, \mathcal{F}_i, \mathcal{J}_i, \mathcal{L}_i\}$, where $\mathcal{A}_i \in \mathbb{R}^{4 \times 3}$ represents the 6D pose of the part in the 3D space, $\mathcal{B}_i \in \mathbb{R}^{4\times 2}$ represents the part's bounding box (BBOX), $\mathcal{G}_i \in \mathbb{R}^{4 \times 2}$ denotes the grasp affordance BBOX, and $\mathcal{F}_i \in \mathbb{R}^{4 \times 2}$ specifies the functional affordance BBOX, $\mathcal{J}_i$ indicates the joint type, and $\mathcal{L}_i$ describes the part's state or function. Notably, for tools that consist of only a single part, the entire object is treated as one part.

To accurately represent each part, we use a rotated BBOX defined by four key points to better fit its geometry and orientation. The $\mathcal{B}$, $\mathcal{G}$ and  $\mathcal{F}$ are defined by their four vertices ${(x_i, y_i)}_{i=0,\ldots,3}$, where each $(x_i, y_i)$ represents the 2D coordinates of \(i\)-th vertex  in the image.

\begin{figure}[t]
    \centering
    \includegraphics[width=1.0\columnwidth]{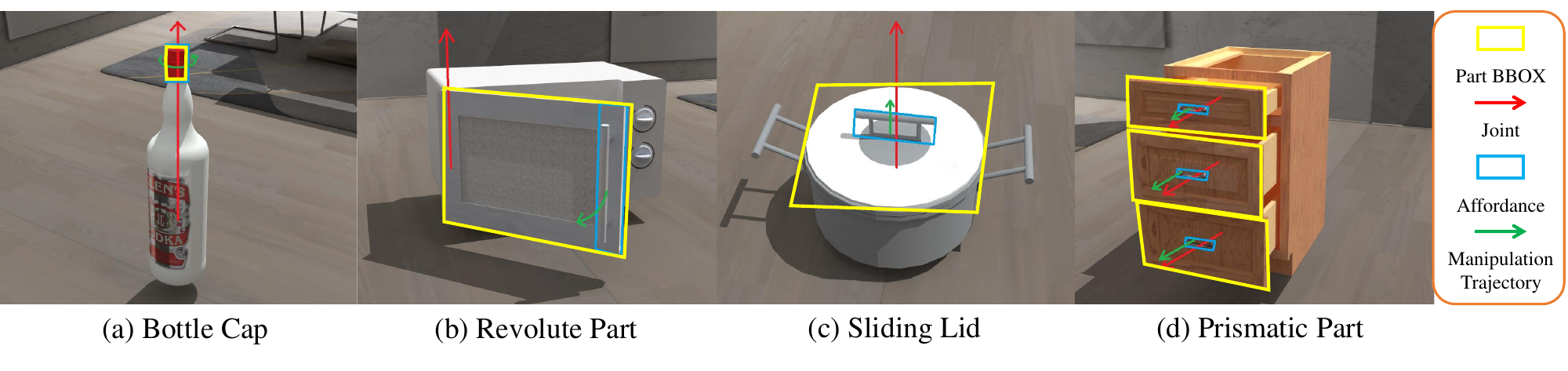}
    \vspace{-0.5cm}
    \caption{Illustration of manipulation types.(a) bottle cap, (b) revolute part, (c) sliding lid, (d) prismatic part. The yellow box represents the object part, the blue box indicates grasp affordance, the red arrow marks the joint parameter, and the green arrow illustrates the manipulation trajectory.}
    \label{Fig: manip_type}
\vspace{-4mm}
\end{figure}

Since most articulated objects consist of one-dimensional prismatic or revolute joints, or a combination of both~\cite{Xiang_2020_SAPIEN}, we categorize $\mathcal{J}$ into four distinct manipulation types based on the manipulation policy, as illustrated in Figure~\ref{Fig: manip_type}, including bottle caps, revolute parts, sliding lids, and prismatic parts, respectively. For tools with 6 DOF, we introduce a fifth manipulation type, termed as the ``freedom object''.

\subsection{Synthetic Data Generation} \label{Subsec: data gen}
Based on the unified structure of part representation formulation $\psi$, which incorporates object-centric 3D spatial motion constraints and corresponding affordances as defined in the previous section, we generate synthetic data in the defined format $\psi$ for tools and articulated objects. Acquiring real-world data is both costly and time-consuming, particularly due to the challenges in obtaining and annotating it. By leveraging near-realistic simulations to generate pre-scanned meshes or URDF models of objects, we efficiently create large-scale data across diverse scenes and object states. This synthetic data enables us to fine-tune VLM models while leveraging the capabilities of large models to achieve generalization from simulation to real-world tasks.

Since the formulation of $\mathcal{B}$, $\mathcal{L}$ remains consistent, we present a unified representation here and will not reiterate it below.
\begin{itemize}
    \item \textit{\textbf{Part BBOX ($\mathcal{B}$)}}: 
    A 2D BBOX defined by four vertices ${(x_i, y_i)}_{i=0,\ldots,3}$ is used to delineate the region of the part in the image.
    \item \textit{\textbf{Descriptive Sentence ($\mathcal{L}$)}}: 
   An explicit description of the part's state or function is essential for VLMs to interpret it effectively.
\end{itemize}

\begin{figure}[t]
    \centering
    \includegraphics[width=0.9\columnwidth]{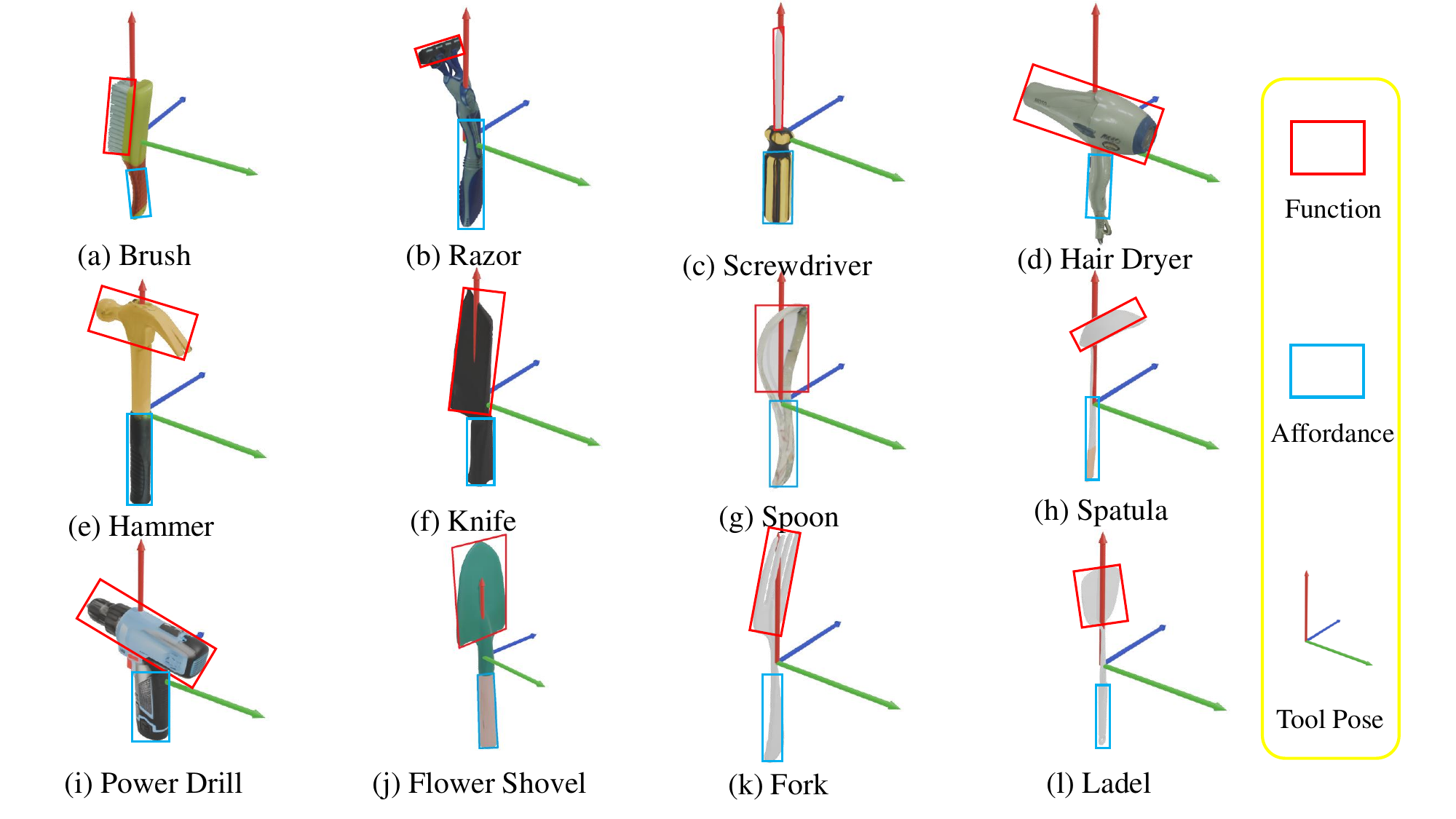}
    \vspace{-0.2cm}
    \caption{Illustration of tools. The blue box indicates grasp affordance, the red box indicates functional affordance and the orientation axis illustrates the object's pose.}
    \label{Fig: tool_type}
\vspace{-2mm}
\end{figure}

\begin{table*}[!ht]
\caption{Overview of the structured manipulation tasks.}
\vspace{-3mm}
\centering
\resizebox{\textwidth}{!}{%
\begin{tabular}{c|c|l|c}
\hline \hline

\textbf{Capabilities} &
\textbf{Tasks} &
\multicolumn{1}{c|}{\textbf{Examples of Task Templates}} &
\textbf{Num.} \\ \hline 

\multicolumn{1}{l|}{Part Grounding.} &
\multicolumn{1}{l|}{2D-Part-Detection} &
\begin{tabular}{@{}l@{}}
    \textcolor{blue}{\texttt{User}}: Please detect all manipulable parts and provide their 2D rotated bounding boxes. \\ 
    \textcolor{teal}{\texttt{UniAff}}:  There are $\mathbf{\mathcal{N}}$ manipulable object parts, each with a corresponding 2D bounding box: $\mathbf{\mathcal{B}_1}$, $\mathbf{\mathcal{B}_2}$, ..., $\mathbf{\mathcal{B}_N}$.
\end{tabular} &

130K \\ \hline 

\multicolumn{1}{l|}{6D Pose and Manipulation Type Understand.} &
\multicolumn{1}{l|}{6D-Pose-Detection} &
\begin{tabular}{@{}l@{}}
    \textcolor{blue}{\texttt{User}}: Please detect the manipulation type of the object and provide the 6D pose. \\ 
    \textcolor{teal}{\texttt{UniAff}}:  \textbf{Manipulation Type} $\mathbf{\mathcal{J}}$ and its \textbf{6D Pose} $\mathbf{\mathcal{A}}$.
\end{tabular} &
130K \\ \hline 

\multicolumn{1}{l|}{Grasp-Affordance Understand.} &
\multicolumn{1}{l|}{2D-Grasp-Affordance} &
\begin{tabular}{@{}l@{}}
     \textcolor{blue}{\texttt{User}}: Please detect the grasp region of part with \textbf{BBox} $\mathbf{\mathcal{B}}$ and provide the 2D bounding box. \\ 
     \textcolor{teal}{\texttt{UniAff}}: Here is the grasp affordance region \textbf{BBox} $\mathbf{\mathcal{G}}$.
\end{tabular} &

130K \\ \hline 

\multicolumn{1}{l|}{Functional-Affordance Understand.} &
\multicolumn{1}{l|}{2D-Functional-Affordance} &
\begin{tabular}{@{}l@{}}
    \textcolor{blue}{\texttt{User}}: Please detect the functional region of part with \textbf{BBox} $\mathbf{\mathcal{B}}$ and provide the 2D bounding box. \\ 
    \textcolor{teal}{\texttt{UniAff}}: Here is the functional affordance region \textbf{BBox} $\mathbf{\mathcal{F}}$.
\end{tabular} &
    
10K \\ \hline \hline

\end{tabular}%
}
\label{tab:VQA_Task}
\vspace{-4mm}
\end{table*}

\subsubsection{\textbf{Tools}}
Modeling a tool’s 6D pose, along with its grasp and functional affordances, is essential for precise and effective manipulation, as the use of a tool is strongly correlated with its 6D pose in 3D space~\cite{li2024learning}. In our work, we utilized 230 tools from real-world scanned data sourced from the PACE~\cite{you2023pace} and OmniObject3D~\cite{wu2023omniobject3d} datasets, supplemented by 370 additional tools that we scanned ourselves, covering 12 categories as shown in Figure~\ref{Fig: tool_type}: brush, razor, screwdriver, hair dryer, hammer, knife, spoon, spatula, power drill, flower shovel, fork, and ladle. To ensure consistency, all tools were aligned to a common axis. Using Blender, we segmented the textured meshes into \textbf{grasp affordance parts} and \textbf{functional affordance parts}, with each tool’s segmentation process taking approximately \textbf{5 minutes}.

Building on these labels and leveraging near-realistic rendering technology in simulation~\cite{Xiang_2020_SAPIEN}, we rendered diverse cluttered scenes. For more details on data rendering, please refer to our~\href{https://sites.google.com/view/uni-aff/home}{\textbf{website}}. 
Each scene was automatically labeled with key attributes defined in the formulation  $\{\mathcal{A}, \mathcal{B}, \mathcal{G}, \mathcal{F}, \mathcal{J}, \mathcal{L}\}$. The manipulation type of the tools is identified as the ``freedom object''. 
\begin{itemize}
    \item \textit{\textbf{6D Pose ($\mathcal{A}$)}}: 
    The $\mathcal{A}$ is represented as a $4 \times 3$ matrix, where the first $1 \times 3$ row represents the tool's 3D spatial position, capturing the translational degrees, and the remaining $3 \times 3$ matrix captures its rotational degrees in 3D.
    \item \textit{\textbf{Grasp Affordances ($\mathcal{G}$)}}: 
    A 2D BBOX defined by four vertices ${(x_i, y_i)}_{i=0,\ldots,3}$ is used to delineate the grasp area of the tool.
    \item \textit{\textbf{Functional Affordances ($\mathcal{F}$)}}:
    A 2D BBOX defined by four vertices ${(x_i, y_i)}_{i=0,\ldots,3}$ is used to represent the functional area of the tool.
\end{itemize}

By varying the scene configurations, tool orientations, and surrounding objects, we ensured that the dataset encompassed a wide range of scenarios, enabling the model to generalize effectively to real-world manipulation tasks.
\subsubsection{\textbf{Articulated Objects}}
To manipulate articulated objects effectively, it is crucial to develop a comprehensive understanding of each part's manipulation type ($\mathcal{J}$), its corresponding joint axis ($\mathcal{A}$) in 3D space, and the grasp affordances ($\mathcal{G}$) associated with various object states. 

To establish a unified representation structure, we relabeled 900 objects across 19 categories from the PartNet-Mobility Dataset~\cite{Xiang_2020_SAPIEN}, including bottle, box, bucket, dispenser, door, folding chair, kitchen pot, laptop, microwave, refrigerator, safe, storage furniture, trash can, faucet, oven, table, toilet, kettle, and washing machine. Since the handles of some objects are not separate meshes or links, we modified the meshes or links of these objects to separate the handles into individual links. This modification allows for more accurate grasp affordance annotation, better aligning with our specific formulation $\psi$.

Based on the definition of $\psi$ in our task formulation, we rendered objects in diverse states. For more details on data rendering, please refer to our~\href{https://sites.google.com/view/uni-aff/home}{\textbf{website}}. We present the detailed auto-labeled dataset as follows:
\begin{itemize}
    \item \textit{\textbf{Manipulation Type ($\mathcal{J}$)}}:
    Each object's movable parts are classified based on their specific 3D motion properties, with joint types divided into four manipulation types ($\mathcal{J}$): bottle caps, sliding lids, revolute parts, and prismatic parts, as illustrated in Figure~\ref{Fig: manip_type}. 
    \item \textit{\textbf{Joint Axis ($\mathcal{A}$)}}: 
    For articulated parts with 4 degrees of freedom (DOF), the joint axis ($\mathcal{A}$) is defined by two points ${(x_i, y_i, z_i)}_{i=0,1}$, representing the joint's axis in 3D space, while the remaining DOF components are masked.
    \item \textit{\textbf{Grasp Affordances ($\mathcal{G}$)}}: 
    An articulated object's affordances change based on its state. For example, a closed door requires the handle to open, while an open door can be manipulated by the handle or edge. To account for this, we use sampling weights that prioritize the door's edge based on the joint state. The sampling weight is calculated as: \begin{equation} weight = \frac{ \mathbf{J} - \theta }{180 - \theta },
    \end{equation}
    where $\mathbf{J}$ represents the current joint state and $\theta$ is a predefined threshold value.
\end{itemize}

\subsubsection{\textbf{VQA Design}}
In our task formulation, we define six key parameters to describe each part: 6D Pose ($\mathcal{A}$), Bounding Box ($\mathcal{B}$), Grasp Affordance Region ($\mathcal{G}$), Functional Affordance Region ($\mathcal{F}$), Manipulation Type ($\mathcal{J}$), and a Sentence Describing the Part’s State or Function ($\mathcal{L}$).

To effectively address these six parameters, we utilize multiple Visual Question Answering (VQA) prompts to design four distinct tasks, as outlined in Table~\ref{tab:VQA_Task}. By guiding the model through a series of task-specific prompts, we develop a multi-task learning framework. By adopting this VQA-based approach, we integrate diverse manipulation-related prior knowledge into a unified model, enabling it to handle complex manipulation tasks across a wide range of object categories and scenarios. These multi-task MLLMs not only improve task efficiency but also enhance the model’s capacity to generalize across diverse real-world robotic manipulation tasks.

\begin{itemize}
    \item \textit{\textbf{2D-Part-Detection Task}}: This task improves the model's ability to identify object parts and their corresponding bounding boxes ($\mathcal{B}$).
    
    \item \textit{\textbf{6D-Pose-Detection Task}}: This task enables the model to detect both the manipulation type ($\mathcal{J}$) and the 6D pose ($\mathcal{A}$) of each part. Accurately determining the joint axis of articulated parts or the 6D pose of tools is essential for guiding the robot’s interactions with them, ensuring precise and efficient task execution. 
    \item \textit{\textbf{2D-Grasp-Affordance Task}}: This task enhances the model's ability to recognize the grasp region of the specified part ($\mathcal{G}$). Accurately identifying these regions is crucial for the robot to effectively grasp the target object and successfully complete the task.
    \item \textit{\textbf{2D-Functional-Affordance Task}}: This task equips the model with the ability to recognize the functional region of the specified part ($\mathcal{F}$), enabling accurate tool usage to complete tasks effectively.

\end{itemize}

\subsection{MLLMs-based Manipulation and Model Fine-tuning} \label{Subsec: model training}

We propose a novel robot manipulation algorithm, termed UniAff, as illustrated in Figure~\ref{Fig: pipeline}. Our model architecture is built upon SPHINX~\cite{lin2023sphinx} and utilizes LLaMA2~\cite{touvron2023llama} as the language backbone,  enabling robust multimodal interaction between visual and linguistic inputs. The design is optimized for fine-grained visual analysis, focusing on capturing detailed regional object features critical for manipulation tasks.

Using the ``Any Resolution" strategy from~\cite{lin2023sphinx}, input images with 448 $\times$ 448 resolution are divided into smaller sub-images to preserve fine-grained details. To ensure comprehensive global and local visual grounding, we integrate the visual encoder from CLIP~\cite{radford2021learning} to extract local semantic features. Additionally, DINOv2~\cite{oquab2023dinov2} is incorporated to further enhance the model's capacity for capturing detailed local semantics, while Q-Former~\cite{li2023blip} is employed to summarize global visual information. The local and global visual features are concatenated along the channel dimension to ensure thorough feature integration and improve visual understanding in manipulation tasks.

Our fine-tuning strategy encodes affordances and 3D physical information within natural language representations, aligning training samples with the VQA framework. Consequently, we adopt cross-entropy loss as our primary training objective. To preserve the model's broad visual reasoning capabilities, we incorporate general visual reasoning tasks alongside those specifically focused on predicting robotic affordances. The visual projection layers and the language model are jointly fine-tuned to ensure effective alignment between visual and linguistic modalities for affordance-based reasoning. Finally, the decoded manipulation information is combined with visual data using a De-Tokenizer,  enabling the completion of the specified task.

\section{Experiments}

In this section, we conduct comprehensive experiments in both simulation and real-world settings. We compare the performance of UniAff with several baseline models to address the following questions: 
1) Can UniAff effectively ground the grasp and functional affordances of tools?
2) Can UniAff simultaneously construct 3D spatial motion constraints along with their corresponding affordances?
3) Can UniAff effectively generalize from perceptual priors to real-world applications?

\subsection{Experimental Settings}

 \noindent{\textbf{Model Setting.}} We fine-tuned the vision-language model based on the SPHINX model~\cite{lin2023sphinx}, utilizing eight NVIDIA A100 GPUs, each with 80 GB of memory. The fine-tuning process was completed over three epochs. To maintain the quality of the pre-trained features, the visual encoders remained frozen throughout the fine-tuning phase. Training was conducted with a batch size of 4, and the learning rate was set to $2 \times 10{^{-5}}$.

 \noindent{\textbf{Dataset Setting.} }
 We selected 9 categories from brush to power drill, creating a training set of 450 tools, including 70 unseen instances. An additional 3 unseen categories include flower shovel, fork, and ladle, totaling 80 tools. We generated 10,000 diverse scenes using the training tools and 3,000 images for unseen instances. We selected 13 categories for articulated objects from bottle to trash can, resulting in a training set of 502 objects and 160 unseen instances. Unseen categories are 6, from faucet to washing machine, totaling 238 objects. Each training object was rendered in 20 states from 5 perspectives, producing a total of 50,200 images. The rendered data were translated according to the structured task formulation $\psi$ to train UniAff, as shown in Table~\ref{tab:VQA_Task}.

\subsection{Robotic Affordance Detection Result}
\noindent{\textbf{Baselines and Metrics.}}
Our robotic affordance evaluation leverages the HANDAL dataset~\cite{guo2023handal}, utilizing pixel-wise segmentation AP. While the baseline model in HANDAL detects only whole objects, our UniAff model identifies both complete objects and manipulable affordance regions effectively. However, because our model generates rotated bounding boxes, a direct comparison of bounding box AP with the ground truth data, which uses standard bounding boxes, is not feasible. Instead, we provide a qualitative comparison with visual examples of our model's performance on our website. To facilitate comparison, we convert the rotated bounding boxes into standard ones and employ SAM~\cite{kirillov2023segment} to generate segmentation masks. This enables us to compare our approach with LISA~\cite{lai2024lisa}, which integrates LLM and a SAM decoder, as well as ManipVQA-SAM~\cite{huang2024manipvqa}, where SAM~\cite{kirillov2023segment} converts the bounding box into segmentation masks.

\noindent{\textbf{Results.}}
As shown in Table~\ref{tab:handal_dataset}, UniAff demonstrates competitive performance even in a zero-shot setting associated with SAM~\cite{kirillov2023segment}. Trained solely on the simulation dataset, UniAff outperformed LISA by a significant margin of $11.5\%$ and achieved performance comparable to ManipVQA, with only a $2.2\%$ difference in IOU. Notably, UniAff does not have access to the HANDAL dataset, unlike ManipVQA.

\begin{table}[t]
\caption{Robotic affordance evaluation results on HANDAL Dataset}
\vspace{-0.3cm}
\label{tab:handal_dataset}
\centering
\resizebox{0.99\columnwidth}{!}{%
\renewcommand\arraystretch{1.3}
\begin{threeparttable}
\begin{tabular}{c|cccccccccc}
\toprule
\multirow{3}{*}{~} 
& \multicolumn{1}{c}{\textbf{Ha}}   &\multicolumn{1}{c}{\textbf{Pd}}&\multicolumn{1}{c}{\textbf{Sd}}&\multicolumn{1}{c}{\textbf{La}}&\multicolumn{1}{c}{\textbf{Pan}}&\multicolumn{1}{c}{\textbf{Sp}}&\multicolumn{1}{c}{\textbf{St}}&\multicolumn{1}{c}{\textbf{Ut}}&\multicolumn{1}{c|}{\textbf{Wh}}&\multicolumn{1}{c}{\textbf{AVG}}\\
\hline     
LISA~\cite{lai2024lisa} &0.671 &0.426 &0.624 &0.407	&0.453 &0.578 &0.397 &0.494 &0.494 
 & \multicolumn{1}{|c}{0.505}\\
ManipVQA~\cite{huang2024manipvqa} &\textbf{0.745} &\textbf{0.439}	&\textbf{0.799} &0.620 &0.622	&\textbf{0.646}	&\textbf{0.638}	&\textbf{0.584}	&0.683
& \multicolumn{1}{|c}{\textbf{0.642}}\\
Ours  &0.742 &0.394	&0.747	&\textbf{0.650} &\textbf{0.680}	&0.621 &0.597 &0.435 &\textbf{0.712}	
& \multicolumn{1}{|c}{0.620}\\
\bottomrule
\end{tabular}
\begin{tablenotes}
\item Object abbreviations are listed in sequence: Hammer, Power Drill, Screwdriver, Ladle, Pan, Spatula, Strainer, Utensil, and Whisk. Results are reported as IOU ($\uparrow$).
\end{tablenotes}
\end{threeparttable}
}
\vspace{-4mm}
\end{table}


\subsection{Tool Usage Understanding Evaluation}
\noindent{\textbf{Baselines and Metrics.}} We evaluate UniAff's tool usage understanding on the test splits of our dataset. Unlike the general affordance evaluation on HANDAL, this evaluation requires grounding both grasp affordance (e.g., grasp area) and functional affordance. To achieve a more compact representation, we use rotated bounding boxes as ground truth. We compare our method against ManipVQA.

\noindent{\textbf{Results.}} As shown in Table~\ref{tab:tool_usage}, UniAff excels at detecting both grasp affordances, with a $32.5\%$ improvement, and functional affordances of tools, with a $56.9\%$ improvement in IOU, compared to ManiVQA, which struggles with functional affordance reasoning. Along with the results in Table~\ref{tab:handal_dataset}, UniAff demonstrates superior generalization, likely due to its larger and more diverse dataset.

\begin{table*}[th]
\caption{Articulated Object Manipulation Results}
\vspace{-0.25cm}
\centering
\begin{threeparttable}
\renewcommand\arraystretch{1.1}
\begin{tabular}{c|ccccccc}
\toprule
\multirow{3}{*}{~} & \multicolumn{4}{c|}{Unseen Instances} & \multicolumn{3}{c}{Unseen Categories} \\
& \multicolumn{1}{c}{Bottle Cap}   &\multicolumn{1}{c}{Sliding Lid}&\multicolumn{1}{c}{Revolute Part} & \multicolumn{1}{c|}{Prismatic Part} & \multicolumn{1}{c}{Sliding Lid} & \multicolumn{1}{c}{Revolute Part} & \multicolumn{1}{c}{Prismatic Part} \\
\hline   
Where2Act~\cite{mo2021where2act}  & 0.2034  & 0.1921  & 0.0897 &	\multicolumn{1}{c|}{0.1093}  & 0.1535  & 0.0822  & 0.0861\\
UMPNet~\cite{wu2021vat} & 0.2787 & 0.2535 & 0.3521 & \multicolumn{1}{c|}{0.3000}  & 0.2789 & 0.3201  & 0.3158 \\
A3VLM~\cite{huang2024a3vlm} & 0.4895& 0.5969   & 0.4397 & \multicolumn{1}{c|}{0.5231}  & 0.4535 & 0.4906  & 0.4387\\
UniAff (w/o afford.) & 0.5166 & 0.6321  &0.4506& \multicolumn{1}{c|}{0.5360}  & 0.4820& 0.5341 & 0.4519 \\
UniAff (ours) & \textbf{0.5259} & \textbf{0.6642}  & \textbf{0.5418}  & \multicolumn{1}{c|}{\textbf{0.6001}}  & \textbf{0.5730} & \textbf{0.5913}  & \textbf{0.5064} \\
\bottomrule
\end{tabular}

\begin{tablenotes}
\item \textbf{Unseen instances} consist of data from four object categories utilized in model training. \textbf{Unseen categories} encompass data from completely new categories. Results are reported as success rates ($\uparrow$).
\end{tablenotes}

\end{threeparttable}
\label{tab:manipulation}
\vspace{-4mm}
\end{table*}

\begin{table*}[th]
\caption{Tool usage understanding evaluation results}
\vspace{-0.55cm}
\label{tab:tool_usage}
\begin{center}
\begin{threeparttable}
\resizebox{0.95\textwidth}{!}{%
\renewcommand\arraystretch{1.1}
\begin{tabular}{c|c|ccccccccccccc}
\toprule

\multirow{3}{*}{Task} & \multirow{3}{*}{Model} & \multicolumn{9}{c|}{Unseen Instances} & \multicolumn{3}{c|}{Unseen Categories} \\
& & \multicolumn{1}{c}{\textbf{Sl}}   &\multicolumn{1}{c}{\textbf{Sp}}&\multicolumn{1}{c}{\textbf{Kf}}&\multicolumn{1}{c}{\textbf{Ra}}&\multicolumn{1}{c}{\textbf{Pd}}&\multicolumn{1}{c}{\textbf{Hd}}&\multicolumn{1}{c}{\textbf{Ha}}&\multicolumn{1}{c}{\textbf{Br}}&\multicolumn{1}{c|}{\textbf{Sd}}&\multicolumn{1}{c}{\textbf{Fs}}&\multicolumn{1}{c}{\textbf{Ld}}&\multicolumn{1}{c|}{\textbf{Fr}}&\multicolumn{1}{c}{\textbf{AVG}}\\

\hline 
\multirow{4}{*}{Grasp} & ManipVQA~\cite{huang2024manipvqa} &0.329&0.487&0.449&0.343&0.461&0.455&0.309&0.360&0.534
& \multicolumn{1}{|c}{0.420}&0.304&0.319
& \multicolumn{1}{|c}{0.398}\\
& 2 Points BBox &0.346&0.511&0.463&0.375&0.596&0.487&0.394&0.445&0.556
& \multicolumn{1}{|c}{0.420}&0.290&0.330
& \multicolumn{1}{|c}{0.434}\\
& 224 Resolution &0.263&0.446&0.377&0.333&0.478&0.406&0.357&0.385&0.390
& \multicolumn{1}{|c}{0.319}&0.236&0.298
& \multicolumn{1}{|c}{0.357}\\
& UniAff (ours) &\textbf{0.751}&\textbf{0.692}&\textbf{0.768}&\textbf{0.717}&\textbf{0.758}&\textbf{0.768}&\textbf{0.684}&\textbf{0.740}&\textbf{0.834}
& \multicolumn{1}{|c}{\textbf{0.713}}&\textbf{0.606}&\textbf{0.641}
& \multicolumn{1}{|c}{\textbf{0.723}}\\
\hline
\multirow{4}{*}{Function} & ManipVQA~\cite{huang2024manipvqa} &0.203&0.271&0.234&0.068&0.043&0.303&0.126&0.106&0.058
& \multicolumn{1}{|c}{0.298}&0.169&0.140
& \multicolumn{1}{|c}{0.168}\\
& 2 Points BBox
&0.610&0.690&0.484&0.476&0.624&0.657&0.502&0.544&0.234
& \multicolumn{1}{|c}{0.594}&0.686&0.551
& \multicolumn{1}{|c}{0.554}\\
& 224 Resolution&0.267&0.527&0.413&0.233&0.254&0.491&0.201&0.413&0.219
& \multicolumn{1}{|c}{0.482}&0.349&0.333
& \multicolumn{1}{|c}{0.349}\\
& UniAff (ours) &\textbf{0.756}&\textbf{0.735}&\textbf{0.779}&\textbf{0.652}&\textbf{0.749}&\textbf{0.767}&\textbf{0.732}&\textbf{0.755}&\textbf{0.689}
& \multicolumn{1}{|c}{\textbf{0.780}}&\textbf{0.745}&\textbf{0.703}	
& \multicolumn{1}{|c}{\textbf{0.737}}\\

\bottomrule
\end{tabular}
}
\begin{tablenotes}
\item Object abbreviations are listed in sequence: Spatula, Spoon, Knife, Razor, Power Drill, Hair Dryer, Hammer, Brush, Screwdriver, Flower Shovel, Ladle, and Fork. Results are reported as IOU ($\uparrow$).
\end{tablenotes}
\end{threeparttable}
\end{center}
\vspace{-0.5cm}
\end{table*}

\subsection{Articulation Manipulation Evaluation}
\noindent{\textbf{Task and Metrics.}} We evaluated articulation tasks, including opening bottle caps, sliding lids, and both revolute and prismatic parts on seen categories. For unseen categories, tasks involved manipulating sliding lids and revolute/prismatic parts. The evaluation covered 160 unseen instances from trained categories and 238 objects from new categories. Success was defined as a binary measure, with a joint state change exceeding the threshold $\delta = 0.1$: success = 1($\delta_{change} \geq \delta$).

\noindent {\bf Baselines.} We compare our proposed method with three different baselines under the identical setting: (1)Where2Act~\cite{mo2021where2act} identifies high-actionability manipulation points and generates short-term actions (e.g., pushing, pulling) at each point for interacting with articulated objects. (2) UMPNet~\cite{xu2022universal} infers action sequences for manipulating objects through self-guided exploration and an Arrow-of-Time attribute. We adapted it by replacing suction with a gripper.
(3)A3VLM\cite{huang2024a3vlm} translates object representations into robot actions utilizing simple primitives. We modified it by using a gripper and GraspNet\cite{fang2020graspnet} to generate and select the highest-scoring grasp pose.






\noindent{{\bf Results.}} We present the success rates in Table~\ref{tab:manipulation}. By explicitly modeling the articulation structure, both A3VLM~\cite{huang2024a3vlm} and UniAff demonstrate superior performance. UniAff achieved a $7.07\%$ improvement in success rates for unseen instances and a $9.60\%$ improvement for unseen categories compared to A3VLM.
\vspace{-4mm}
\begin{figure}[H]
\centering\includegraphics[width=0.9\columnwidth]{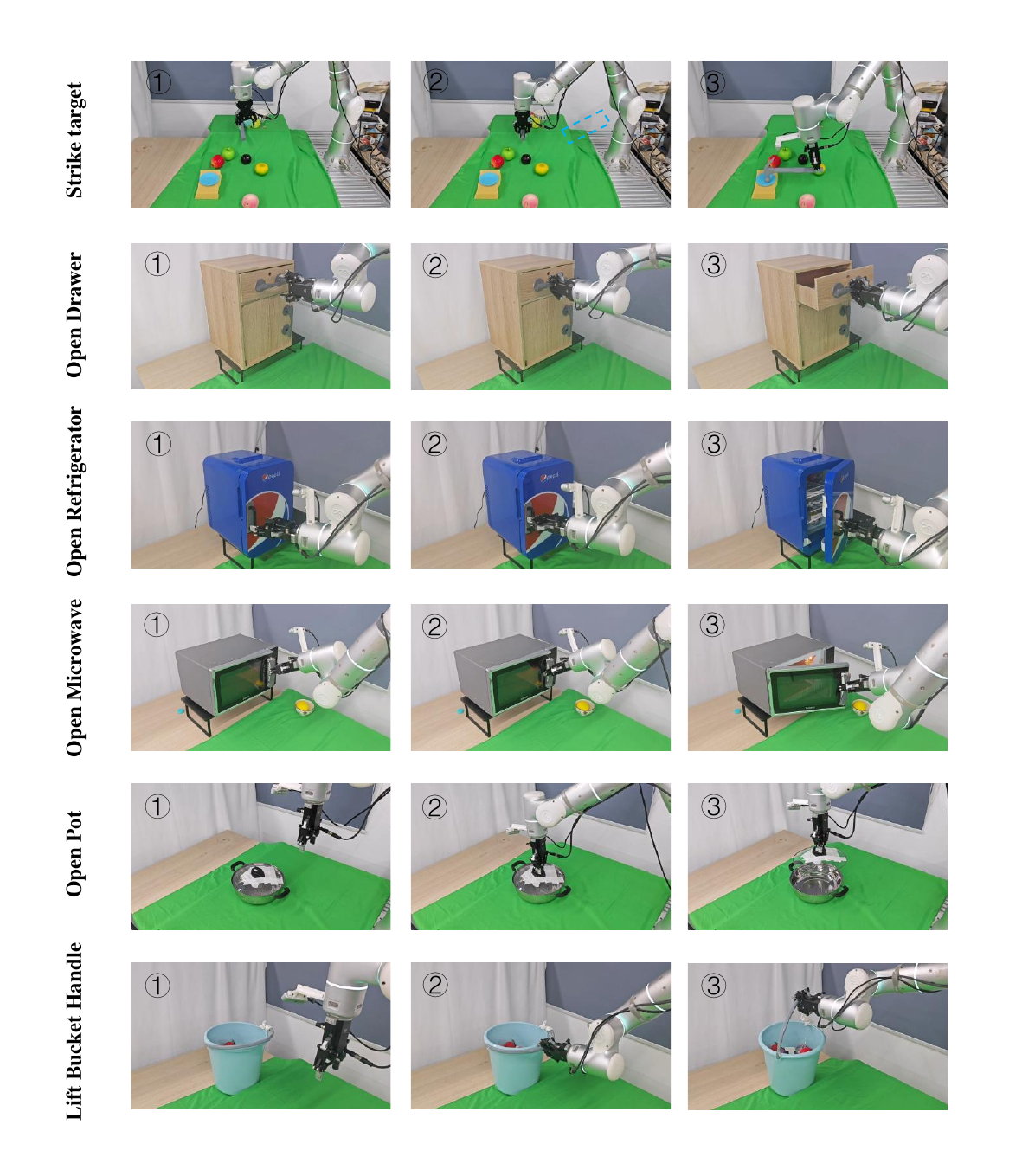}
\vspace{-4mm}
    \caption{Implementation of UniAff in real-world experiments progressed from tool manipulation to articulated object interaction, encompassing tasks such as striking a designated target with a hammer, opening a drawer, refrigerator, microwave, pot, and lifting a bucket handle.}
    \label{Fig:real_world}
\vspace{-2mm}
\end{figure}

\subsection{Ablation Studies}
To assess UniAff's design contributions, we conducted ablation studies. Results in Table~\ref{tab:tool_usage} indicate that omitting the “Any Resolution” method (e.g., using a 224 × 224 resolution) significantly degrades performance, highlighting the necessity of high-resolution input for partial object understanding. Additionally, the use of a 2-point bounding box also reduces performance, as rotated bounding boxes provide a more compact representation for affordance understanding.


\subsection{Real-World Experiments}
We conducted real-world experiments using a 7-DoF Flexiv robot. First, we mounted an RGB-D RealSense D435 camera on the robot’s wrist to capture RGB-D images of various objects. We then applied UniAff to predict object-centric 3D motion constraints and corresponding affordances. The experiments involved six tasks: striking a designated target with a hammer, opening a drawer, opening a refrigerator, opening a microwave, opening a pot, and lifting a bucket handle. These tasks demonstrated UniAff's ability to generalize effectively to real-world manipulation tasks, as illustrated in Figure~\ref{Fig:real_world}. Due to space constraints, detailed information is provided in the~\href{https://sites.google.com/view/uni-aff/home}{\textbf{website}}.

\section{Conclusion}

In this paper, we propose UniAff, a novel approach integrating object-centric 3D motion constraints and affordances for manipulation tasks. By leveraging multimodal large language models (MLLMs), UniAff improves manipulation knowledge and generates precise 3D motion constraints and affordances for diverse objects.  Our extensive dataset labeled with manipulation-related key attributes, comprising 900 articulated objects and 600 tools, serves as a foundation for training UniAff to generalize across a wide range of tasks. Experiments show UniAff significantly outperforms existing methods in both simulation and real-world environments, demonstrating superior generalizability in manipulation tasks involving tools and articulated objects.

{\small
\bibliographystyle{IEEEtranN}
\bibliography{ref}
}

\end{document}


\maketitle
\section{Synthetic Data rendering}
In this section, we will introduce synthetic data rendering related to the structured 3D spatial formulation for object representation in robotic manipulation introduced in \textbf{Method}. First, we will import the pre-labeled articulated objects' URDF or the tool's mesh into the SAPIEN simulator~\cite{Xiang_2020_SAPIEN} and set up the camera, and then we will obtain the corresponding scene image $I$ and depth map $D$. After that, we will determine the corresponding 6D pose $\mathcal{A}$,  part BBOX $\mathcal{B}$, grasp affordance BBOX $\mathcal{G}$, functional affordance BBOX $\mathcal{F}$, joint type $\mathcal{J}$  for each part based on the part-level masks obtained from the simulator.

\clearpage
{\small
\bibliographystyle{IEEEtranN}
\bibliography{ref}
}